\documentclass{article}

\usepackage{amsmath,amsfonts,bm}

\def\eqref#1{equation~\ref{#1}}

\def\1{\bm{1}}

\def\vz{{\bm{z}}}

\def\mH{{\bm{H}}}

\def\mW{{\bm{W}}}
\def\mX{{\bm{X}}}

\DeclareMathAlphabet{\mathsfit}{\encodingdefault}{\sfdefault}{m}{sl}
\SetMathAlphabet{\mathsfit}{bold}{\encodingdefault}{\sfdefault}{bx}{n}

\usepackage{hyperref}
\usepackage{url}

\usepackage{caption}

\usepackage{tabularx}
\usepackage{amsmath,amssymb,amsthm}
\usepackage{graphicx}
\usepackage{subcaption}
\usepackage{listings}
\usepackage{upquote}
\usepackage{wrapfig}
\usepackage{lipsum}
\usepackage{array}
\usepackage{multicol}
\usepackage{multirow}
\usepackage{url}
\usepackage{color}
\usepackage{wrapfig,lipsum,booktabs}
\usepackage{makecell}
\usepackage{booktabs}
\usepackage{changepage}
\usepackage{xcolor}

\usepackage[noend]{algpseudocode}
\usepackage{tabulary}
\newcolumntype{K}[1]{>{\centering\arraybackslash}p{#1}}

\usepackage{tabu}
\usepackage{arydshln}

\usepackage[accepted]{icml2019}

\begin{document}

\twocolumn[
\icmltitle{Improving Sentence Representations with Consensus Maximisation}




\begin{icmlauthorlist}
\icmlauthor{Shuai Tang}{to}
\icmlauthor{Virginia R. de Sa}{to}
\end{icmlauthorlist}

\icmlaffiliation{to}{Department of Cognitive Science, UC San Diego, La Jolla, CA, USA}

\icmlcorrespondingauthor{Shuai Tang}{shuaitang93@ucsd.edu}

\icmlkeywords{Machine Learning, ICML}

\vskip 0.3in
]



\printAffiliationsAndNotice{}  

\begin{abstract}
Consensus maximisation learning can provide self-supervision when different views are available of the same data. The distributional hypothesis provides another form of useful self-supervision from adjacent sentences which are plentiful in large unlabelled corpora. Motivated by the observation that different learning architectures tend to emphasise different aspects of sentence meaning, we present a new self-supervised learning framework for learning sentence representations which minimises the disagreement between two views of the same sentence where one view encodes the sentence with a recurrent neural network (RNN), and the other view encodes the same sentence with a simple linear model. After learning, the individual views (networks) result in higher quality sentence representations than their single-view learnt counterparts (learnt using only the distributional hypothesis) as judged by performance on standard downstream tasks. An ensemble of both views provides even better generalisation on both supervised and unsupervised downstream tasks. Also, importantly the ensemble of views trained with consensus maximisation between the two different architectures performs better on downstream tasks than an analogous ensemble made from the single-view trained counterparts.
\end{abstract}

\section{Introduction}
Consensus maximisation methods provide the ability to extract information from different views of the data and enable self-supervised learning of useful features for future prediction when annotated data is not available \cite{deSa1993LearningCW}. Minimising the disagreement among multiple views helps the model to learn rich feature representations of the data and, also after learning, the ensemble of the feature vectors from multiple views can provide an even stronger generalisation ability.

Distributional hypothesis \cite{harris1954distributional} noted that words that occur in similar contexts tend to have similar meaning \cite{Turney2010FromFT}, and distributional similarity \cite{firth57synopsis} consolidated this idea by stating that the meaning of a word can be inferred by the company it has. The hypothesis has been widely used in the  machine learning community to learn vector representations of human languages. Models built upon distributional similarity don't explicitly require human-annotated training data; the supervision comes from the semantic continuity of language data, such as text and speech.

Large quantities of annotated data are usually hard and costly to obtain, thus it is important to study unsupervised and self-supervised learning. Our goal is to propose learning algorithms built upon the ideas of consensus maximisation learning and the distributional hypothesis to learn from unlabelled data. 

Our proposed consensus maximisation algorithm aims to leverage the functionality of both RNN-based models, which have been widely applied in sentiment analysis tasks \cite{Yang2016HierarchicalAN}, and the linear/log-linear models, which have excelled at capturing attributional similarities of words and sentences \cite{Arora2016ALV,Arora2017ASB,Hill2016LearningDR,Turney2010FromFT} for learning sentence representations. Previous work on unsupervised sentence representation learning based on distributional hypothesis can be roughly categorised into three types:

\textbf{Generative objective:} These models generally follow the encoder-decoder structure. The encoder learns to produce a vector representation for the current input, and the decoder learns to generate sentences in the adjacent context given the produced vector \cite{Kiros2015SkipThoughtV,Hill2016LearningDR,Gan2017LearningGS,Tang2018SpeedingUC}. The idea is straightforward, yet its scalability for very large corpora is hindered by the slow decoding process that dominates training time. Also the decoder in each model is discarded after learning as the quality of generated sequences is not the main concern, which is a waste of parameters and learning effort. 

\textbf{Discriminative Objective:} In these models, a classifier is learnt on top of the encoder/encoders to distinguish adjacent sentences from those that are not \cite{Li2014AMO,Jernite2017DiscourseBasedOF,Nie2017DisSentSR,logeswaran2018an}; these models make a prediction using a predefined differentiable similarity function on the representations of the input sentence pairs or triplets, and the inductive biases the encoders introduce and the objective function have a strong impact on how well the learnt sentence representations generalise to downstream tasks. 

\textbf{No Learning Required:} Instead of learning encoders from large unlabelled corpora, research effort has been put into developing postprocessing methods for constructing sentence vectors from pretrained word vectors \cite{Arora2017ASB,Ethayarajh2018UnsupervisedRW}. The proposed methods excel on sentence-level textual similarity tasks, but not on  single-sentence classification tasks. In contrast, an ensemble of word vectors projected through random linear projections provides solid performance on mostly single-sentence classification tasks, while the effectiveness on the textual similarity tasks is still worth studying and discussing.

Motivated by the discussed prior work, we present a consensus maximisation learning method with a discriminative objective, that uses an RNN encoder and a linear encoder; it learns to maximise agreement between the outputs of the two networks among adjacent sentences. Compared to earlier work on consensus maximisation, especially conventional multi-view learning methods, \cite{deSa1993LearningCW,Dhillon2011MultiViewLO,Wang2015OnDM} that take data from various sources or split data into disjoint populations, our learning method processes the exact \textbf{same} data in two \textbf{distinctive} ways. Since our proposed method lies in the general framework of multi-view learning \cite{RpingLearningWM} with consensus maximisation, the two encoding functions are referred to as two different views in our paper. The two distinctive information processing views tend to encode different aspects of an input sentence; forcing agreement/alignment between these pathways encourages each model to learn a better representation, and is beneficial to the future use of the learnt representations.

Our contribution is threefold:

$\bullet$ A consensus maximisation based method for learning sentence representations is proposed, which adopts a discriminative objective. Two encoding functions, an RNN and a linear model, are learnt jointly in our proposed method.

$\bullet$ The results show that in the proposed method, maximising the agreement between representations from two encoding functions/views gives improved performance of each individual view on all evaluation tasks compared to their single-view trained counterparts, and furthermore ensures that the ensemble of two views provides even better results than each improved view alone.

$\bullet$ Comparisons among variants of our proposed method indicate that maximising the agreement between representations from an RNN and a linear model does more than regularising the same RNN model with a skip connection from the input layer to the output layer.


It is shown \cite{Hill2016LearningDR} that the consistency between supervised and unsupervised evaluation tasks is much lower than that within either supervised or unsupervised evaluation tasks alone and that a model that performs well on supervised evaluation tasks may fail on unsupervised tasks. It is subsequently showed \cite{Conneau2017SupervisedLO, subramanian2018learning} that, with large-scale labelled training corpora, the resulting representations of the sentences from the trained model excel in both supervised and unsupervised tasks, while the labelling process is costly. Our model is able to achieve good results on both groups of tasks {\bf without} labelled information.

\section{Model Architecture}

Our goal is to marry \textbf{RNN-based} sentence encoder and the \textbf{avg-on-word-vectors} sentence encoder into multi-view frameworks with simple objectives. 
The motivation for the idea is that, 
RNN-based encoders process the sentences sequentially, and are able to capture complex syntactic interactions, while the avg-on-word-vectors encoder has been shown to be good at capturing the coarse meaning of a sentence which could be useful for finding paradigmatic parallels \cite{Turney2010FromFT}.

We present our consensus maximisation learning method, and it learns two different sentence encoders jointly; after learning, the vectors produced from two encoders of the same input sentence are used to compose the sentence representation. The details of our proposed learning method is described as follows:

\subsection{Encoders}
In our consensus maximisation learning method, we first introduce two encoders that, after learning, can be used to build sentence representations. One encoder is a bi-directional Gated Recurrent Unit \cite{Chung2014EmpiricalEO} $f(s;\boldsymbol{\phi})$, where $s$ is the input sentence and $\boldsymbol{\phi}$ is the parameter vector in the GRU. During learning, only hidden state at the last time step is fed to the next stage in learning. The other encoder is a linear avg-on-word-vectors model $g(s;\mW)$, which basically transforms word vectors in a sentence by a learnable weight matrix $\mW$ and outputs an averaged vector.

\subsection{Discriminative Objective}
The discriminative objective learns to maximise the agreement between the representations of a sentence pair across two views if one sentence in the pair is in the neighbourhood of the other one. An RNN encoder $f(s;\boldsymbol{\phi})$ and a linear avg-on-word-vectors $g(s;\mW)$ produce a vector representation $\vz^f_i$ and $\vz^g_i$ for $i$-th sentence respectively. The agreement between two views of a sentence pair $(s_i,s_j)$ is defined as $a_{ij}=a_{ji}=\cos(\vz_i^f,\vz_j^g)+\cos(\vz_i^g,\vz_j^f)$. The training objective is to minimise the loss function:
\begin{align}
\mathcal{L}(\boldsymbol{\phi},\mW)&=-\sum_{|i-j|\leq c}\log p_{ij},  \label{dis_loss} \\
\text{ where } p_{ij} &= \frac{e^{a_{ij}/\tau}}{\sum_{n=i-N/2}^{i+N/2-1}e^{a_{in}/\tau}}
\end{align}
where $\tau$ is the learnable temperature term, which is essential for exaggerating the difference between adjacent sentences and those that are not. The neighbourhood/context window $c$, and the batch size $N$ are hyperparameters. 

The choice of cosine similarity based agreement between two views is based on the observations \cite{Turney2010FromFT} that, of word vectors derived from distributional similarity, vector length tends to correlate with frequency of words, thus angular distance captures more important meaning-related information. Also, since our model is unsupervised/self-supervised, whatever similarity there is between neighbouring sentences is what is learnt as important for meaning.

\subsection{Postprocessing}
The postprocessing step \cite{Arora2017ASB}, which removes the top principal component of a batch of representations, is applied on produced representations from $f$ and $g$ respectively after learning with a final $l_2$ normalisation. 

In our consensus maximisation learning method with discriminative objective, in order to reduce the discrepancy between training and testing, the top principal component is estimated by the \textbf{power iteration} method \cite{mises1929praktische} and removed during learning.

\section{Experimental Design}
The proposed method along with its variants are trained on three unlabelled corpora from different genres individually, including BookCorpus \cite{Zhu2015AligningBA}, UMBC News \cite{han2013umbc_ebiquity} and Amazon Book Review \footnote{Largest subset of Amazon Review.}\cite{McAuley2015ImageBasedRO}; The summary statistics of the three corpora can be found in Table \ref{stats}. Adam optimiser \cite{Kingma2014AdamAM} and gradient clipping \cite{Pascanu2013OnTD} are applied for stable training. Pretrained word vectors, fastText \cite{Bojanowski2017EnrichingWV}, are used in our frameworks and fixed during learning. After learning, the parameters in each model are frozen and the learnt model is used as a general purpose sentence representation extractor on downstream NLP tasks. The overall performance on downstream tasks indicates how well the produced representations from the learnt model generalise.

\begin{table}[!ht]
\fontsize{8}{10}\selectfont
\caption{\textbf{Summary statistics} of the three corpora used in our experiments. For simplicity, the three corpora will be referred to as \textbf{D1}, \textbf{D2} and \textbf{D3} in the following tables respectively.}
\begin{center}
    \vskip 0.15in
    \tabulinesep =_2pt^3pt
    \begin{tabu}to \textwidth{@{}c|c|c@{}}
    \toprule
    \multirow{2}{*}{Name} & \multirow{2}{*}{\# of sentences} & mean \# of words  \\
                        & &     per sentence \\
    \midrule
    BookCorpus (\textbf{D1}) & 74M & 13 \\
    UMBC News (\textbf{D2}) & 134.5M & 25 \\
    Amazon Book Review (\textbf{D3}) & 150.8M & 19 \\
    \bottomrule
\end{tabu}
\end{center}
\label{stats}
\vskip -0.1in
\end{table}

\begin{table}[!ht]
    \caption{\textbf{Representation pooling in testing phase.} ``max($\cdot$)'', ``avg($\cdot$)'', and ``min($\cdot$)'' refer to global max-, mean-, and min-pooling over time, which result in a single vector. The table also presents the diversity of the way that a single sentence representation can be calculated. $\mX_i$ refers to word vectors in $i$-th sentence, and $\mH_i$ refers to hidden states at all time steps produced by $f$.} 
    \fontsize{8}{10}\selectfont
    \centering
    \vskip 0.15in
    \tabulinesep =_3pt^3pt
    \begin{tabu}to \textwidth{@{}c||c|c@{}}
    \midrule
        \multirow{2}{*}{Phase} & \multicolumn{2}{c}{Testing} \\
        \cline{2-3}
         &   Supervised & Unsupervised  \\
         \hline
        \midrule
        $f$: $\vz^f_i$ & $[\text{max}(\mH_i);\text{avg}(\mH_i);\text{min}(\mH_i);\mathbf{h}_i^{M_i}]$ & $\text{avg}(\mH_i)$\\
        \midrule
        $g$: $\vz^g_i$ & $[\text{max}(\mW\mX_i);\text{avg}(\mW\mX_i);\text{min}(\mW\mX_i)]$ & $\text{avg}(\mW\mathbf{X}_i)$\\
        \midrule
        Ensemble & Concatenation & Averaging \\
        \midrule
    \end{tabu}
    \label{usage}
\end{table}

All of our experiments including training and testing are done in PyTorch \cite{paszke2017automatic}. The modified SentEval \cite{Conneau2018SentEvalAE} package with the step that removes the first principal component is used to evaluate our learning method on the downstream tasks. Hyperparameters, including batch size $N$ and context window $c$ are tuned only on the averaged performance on STS14 of the model trained on the BookCorpus; STS14/D1 results are thus marked with a $\star$ in Table \ref{unsupervised} and Table \ref{FS_QT} to indicate possible overfitting on that dataset/model only. Batch size $N$ and dimension $d$ in our proposed learning method and its variants are set to be the same for fair comparison. 

\begin{table*}[!ht]
\caption{\textbf{Model study.} Variants of our consensus maximisation learning method with two distinctive views are tested to illustrate the advantage of our learning with two different views. \textbf{In general, with the proposed learning method, learning to maximise the agreement between representations from both views helps each view to perform better and an ensemble of both views provides stronger results than each of them.} The arrow and value pair indicate how a result differs from our consensus maximisation learning method. Better view in colour. }
\fontsize{8}{10}\selectfont
\begin{center}
\vskip 0.15in
\tabulinesep =_3pt^3pt
\begin{tabu}to \textwidth{@{} c | c | c | c | c | c @{}}
        \hline

        & & Unsupervised tasks & \multicolumn{3}{c}{Supervised tasks} \\
        \cline{3-6}
        \textbf{UMBC} & \multirow{3}{*}{Hrs} & Avg of STS tasks  & Avg of STS tasks & Avg of Binary-CLS tasks & \multirow{2}{*}{MRPC} \\        
        \textbf{News} & & (STS12-16, SICK14) & (SICK-R, STS-B) & (MR, CR, SUBJ, MPQA, SST) &  \\
        \Xhline{2pt}
        \hline
        \multicolumn{6}{c}{\textbf{Our Multi-view with Discriminative Objective}: $a_{ij}=\cos(\vz_i^f,\vz_j^g)+\cos(\vz_i^g,\vz_j^f)$} \\[0.5ex]

        \hline
        $\vz^f$ & \multirow{3}{*}{8} & 67.4 & 83.0 & 86.6 & 75.5/82.7\\
        $\vz^g$ & & 69.2 & 82.6 & 85.2 & 74.3/82.7 \\
        $\text{ensemble}(\vz^f, \vz^g)$ & & 71.4 & 83.0 & 86.6 & 76.8/83.9 \\
    
        \Xhline{2pt}
        \multicolumn{6}{c}{Multi-view with $f_1$ and $f_2$: $a_{ij}=\cos(\vz_i^{f_1},\vz_j^{f_2})+\cos(\vz_i^{f_2},\vz_j^{f_1})$} \\[0.5ex]
        \multicolumn{6}{c}{Multi-view with $g_1$ and $g_2$: $a_{ij}=\cos(\vz_i^{g_1},\vz_j^{g_2})+\cos(\vz_i^{g_2},\vz_j^{g_1})$} \\[0.5ex]
        \hline
        $\vz^{f_1}$ & \multirow{2}{*}{17} & 49.7 \textcolor{red}{($\downarrow$17.7)} & 82.2 \textcolor{red}{($\downarrow$0.8)} & 86.3 \textcolor{red}{($\downarrow$0.3)} & 75.9/83.0  \\
        $\text{emsemble}(\vz^{f_1}, \vz^{f_2})$ & & 57.3 \textcolor{red}{($\downarrow$14.1)} & 81.9 \textcolor{red}{($\downarrow$1.1)} & 87.1 \textcolor{green}{($\uparrow$0.5)} & 77.2/83.7 \\

        \hline
        $\vz^{g_1}$ & \multirow{2}{*}{2} & 68.5 \textcolor{red}{($\downarrow$0.7)} & 80.8 \textcolor{red}{($\downarrow$1.8)} & 84.2 \textcolor{red}{($\downarrow$1.0)} & 72.5/82.0 \\
        $\text{emsemble}(\vz^{g_1}, \vz^{g_2})$ & & 69.1 \textcolor{red}{($\downarrow$2.3)} & 77.0 \textcolor{red}{($\downarrow$6.0)} & 84.5 \textcolor{red}{($\downarrow$2.1)} & 73.5/82.3\\

        \hline
        $\text{emsemble}(\vz^{f_1}, \vz^{g_1})$ & 19 & 67.5 \textcolor{red}{($\downarrow$3.9)} & 82.3 \textcolor{red}{($\downarrow$0.7)} & 86.9 \textcolor{green}{($\uparrow$0.3)} & 76.6/83.8 \\

        \Xhline{2pt}

        \multicolumn{6}{c}{Single-view with $f$ only: $a_{ij}=\cos(\vz_i^f,\vz_j^f)$, Single-view with $g$ only: $a_{ij}=\cos(\vz_i^g,\vz_j^g)$} \\[0.5ex]

        \hline
        $\vz^f$ & 9 & 57.8 \textcolor{red}{($\downarrow$9.6)} & 81.6 \textcolor{red}{($\downarrow$1.4)} & 85.8 \textcolor{red}{($\downarrow$0.8)} & 74.8/82.3 \\
        $\vz^g$ & 1.5 & 68.7 \textcolor{red}{($\downarrow$0.5)} & 81.1 \textcolor{red}{($\downarrow$1.5)} & 83.3 \textcolor{red}{($\downarrow$1.9)} & 72.9/81.0 \\
        $\text{emsemble}(\vz^f, \vz^g)$ & 10.5 & 68.6 \textcolor{red}{($\downarrow$2.8)} & 82.3 \textcolor{red}{($\downarrow$0.7)} & 86.3 \textcolor{red}{($\downarrow$0.3)} & 75.4/82.5 \\
        \hline

        \Xhline{2pt}
        \multicolumn{6}{c}{Multi-view + Single-view with $f$ and $g$: $a_{ij}=\cos(\mathbf{z}_i^f,\mathbf{z}_j^g)+\cos(\mathbf{z}_i^g,\mathbf{z}_j^f)+\cos(\mathbf{z}_i^f,\mathbf{z}_j^f)+\cos(\mathbf{z}_i^g,\mathbf{z}_j^g)$} \\[0.5ex]
        \hline
        $\mathbf{z}^f$ &                    & 66.3 \textcolor{red}{($\downarrow$1.1)} & 82.1 \textcolor{red}{($\downarrow$0.9)} & 85.9 \textcolor{red}{($\downarrow$0.7)} & 75.2/83.4 \\
        $\mathbf{z}^g$ & \multirow{3}{*}{8} & 70.2 \textcolor{green}{($\uparrow$1.0)} & 81.8 \textcolor{red}{($\downarrow$0.8)} & 84.3 \textcolor{red}{($\downarrow$0.9)} & 72.6/81.3 \\
        $\text{emsemble}(\mathbf{z}^f, \mathbf{z}^g)$ &  & 69.9 \textcolor{red}{($\downarrow$2.4)} & 82.7 \textcolor{red}{($\downarrow$0.3)} & 86.5 \textcolor{red}{($\downarrow$0.1)} & 75.8/82.9 \\
        \hline
        \Xhline{2pt}
        \multicolumn{6}{c}{Skip Connection as Regularisation with $f$ and $g$: $a_{ij}=\cos(\vz_i^f + \vz_i^g, \vz_j^f + \vz_j^g)$} \\[0.5ex]
        \hline 
        $\vz^f$ & \multirow{3}{*}{8}    & 59.4 \textcolor{red}{($\downarrow$8.0)}  & 81.4 \textcolor{red}{($\downarrow$1.6)}        & 85.7 \textcolor{red}{($\downarrow$0.9)} & 75.1/83.0 \\        
        $\vz^g$                       & & 52.8 \textcolor{red}{($\downarrow$16.4)} & 81.3 \textcolor{red}{($\downarrow$1.3)}        & 84.1 \textcolor{red}{($\downarrow$1.1)} & 73.3/81.7 \\
        $\text{emsemble}(\vz^f, \vz^g)$            & & 64.5 \textcolor{red}{($\downarrow$6.9)}  & 83.0 \textcolor{blue}{(-)}                     & 86.4 \textcolor{red}{($\downarrow$0.2)} & 76.2/83.8 \\
        \hline
    \end{tabu}
\end{center}
\label{UMBCmvsv}
\vskip -0.1in
\end{table*}

\subsection{Unsupervised Evaluation - Textual Similarity}
\textbf{Representation:} For a given sentence input $s$ with $M$ words, suggested by \cite{Pennington2014GloveGV, Levy2015ImprovingDS}, the representation is calculated as $\vz=\left(\hat{\vz}^f+\hat{\vz}^g\right)/2$, where $\hat{\vz}$ refers to the post-processed and normalised vector, and is mentioned in Table \ref{usage}.

\textbf{Tasks}: The unsupervised tasks include five tasks from SemEval Semantic Textual Similarity (STS) in 2012-2016 \cite{Agirre2012SemEval2012T6,Agirre2013SEM2S,Agirre2014SemEval2014T1,Agirre2015SemEval2015T2,Agirre2016SemEval2016T1} and the Semantic Relatedness task (SICK-R) \cite{Marelli2014ASC}. 

\subsection{Supervised Evaluation}
The evaluation on these tasks involves learning a linear model on top of the learnt sentence representations produced by the model. Since a linear model is capable of selecting the most relevant dimensions in the feature vectors to make predictions, it is preferred to concatenate various types of representations to form a richer, and possibly more redundant feature vector, which allows the linear model to explore the combination of different aspects of encoding functions to provide better results.

\textbf{Representation:}  Inspired by prior work \cite{McCann2017LearnedIT,Shen2018BaselineNM}, the representation $\vz^f$ is calculated by concatenating the outputs from the global mean-, max- and min-pooling on top of the hidden states $\mH$, and the last hidden state, and $\vz^g$ is calculated with three pooling functions as well. The post-processing and the normalisation step is applied individually. These two representations are concatenated to form a final sentence representation. Table \ref{usage} presents the details.

\textbf{Tasks}: Semantic relatedness (SICK) \cite{Marelli2014ASC}, paraphrase detection (MRPC) \cite{Dolan2004UnsupervisedCO}, question-type classification (TREC) \cite{Li2002LearningQC}, movie review sentiment (MR) \cite{Pang2005SeeingSE}, Stanford Sentiment Treebank (SST) \cite{Socher2013RecursiveDM}, customer product reviews (CR) \cite{Hu2004MiningAS}, subjectivity/objectivity classification (SUBJ) \cite{Pang2004ASE}, opinion polarity (MPQA) \cite{Wiebe2005AnnotatingEO}. 

\section{Model Study}
Since our proposed learning method is based on the consensus maximisation learning with two distinctive information processing views, it is important to study why our certain configuration is superior than its other variants, and to understand how two views interact with each other. Table \ref{UMBCmvsv} presents the macro-averaged results on downstream tasks of our proposed methods and its reasonable comparison variants/counterparts.\footnote{The unsupervised training is conducted on UMBC News as BookCorpus is not publicly avaiable anymore.}

\subsection{Consensus Maximisation with Same Architecture}
In order to determine if the consensus maximisation learning method with two different views/encoding functions, $f$ and $g$, is helping the learning, two models, each with two encoding functions of the same type but parametrised independently, either two $f$-s or two $g$-s, are trained on the same large corpus and then evaluated on downstream tasks. The agreement between two views for each model is defined: 
\begin{align}
\text{$f_1$ and $f_2$}: & a_{ij}=\cos(\vz_i^{f_1},\vz_j^{f_2})+\cos(\vz_i^{f_2},\vz_j^{f_1}) \\
\text{$g_1$ and $g_2$}: & a_{ij}=\cos(\vz_i^{g_1},\vz_j^{g_2})+\cos(\vz_i^{g_2},\vz_j^{g_1}) 
\end{align}
As specifically emphasised in prior work \cite{Hill2016LearningDR}, linear/log-linear models, which include $g$ in our model, produce better representations for unsupervised evaluation tasks than RNN-based models do. This can be observed in Table \ref{UMBCmvsv} as well, where $g$ consistently provides better results on unsupervised tasks than $f$. 

It is shown in the second section in Table \ref{UMBCmvsv} that \textbf{both $\vz^f$ and $\vz^g$ gets improved in our learning method} in a way that $f$ gives better performance on unsupervised evaluation tasks and $g$ does better on supervised tasks after learning to maximising the agreement between two views, compared to the models trained with only one type of network architecture. This observation supports our hypothesis that two different views are helping each other to learn more generalisable representations, and it also distinguishes our consensus maximisation learning method from other conventional multi-view learning algorithms that collect data from multiple domains or split data into multiple disjoint populations. Our results show that consensus maximisation learning 
can benefit not only from two conventional views of the data but also from two constructed views that have different  properties, neural network architectures, or information processing methods. 

The ensemble of two independently trained $\vz^f$ and $\vz^g$ provides worse results than $\vz^g$ does solely on unsupervised evaluation tasks, while in our jointly learnt method, \textbf{the ensemble of $\vz^f$ and $\vz^g$ performs even better than each one of them}. It suggests that maximising representations between two views ensures that the ensemble of them give better results than each of them on unsupervised tasks, and according to Table \ref{UMBCmvsv}, it provides more performance gain on supervised evaluation tasks than learning two views independently does.

\subsection{Single-view learning}
A similar comparison counterpart is to learn a single model, either $f$ or $g$, to produce more similar representations for adjacent sentences than those that are not. Given the discussion above, we can expect that our learning method with two jointly learnt views outperform their single-view learnt variants, and the ensemble perform of two views performs even better. Results in the third section in Table \ref{UMBCmvsv} meet our expectations. The agreement in the single-view learning method is calculated as follows:
\begin{align}
f: a_{ij}=\cos(\vz_i^f,\vz_j^f)  \text{ \hspace{0.1in}   and    \hspace{0.1in}   }  g: a_{ij}=\cos(\vz_i^g,\vz_j^g)
\end{align}
Interestingly, compared to a model with two views but of the same type, the single-view learnt model doesn't provide worse performance overall, which may suggest that increasing parameters blatantly wouldn't lead to better evaluation results in unsupervised learning, however, a proper inductive bias, such as the maximising the agreement between two different encoding functions would empower the generalisation ability of the learnt representations. 

\begin{figure}[!ht]
\vskip 0.2in
    \begin{center}
    \includegraphics[width=0.45\textwidth]{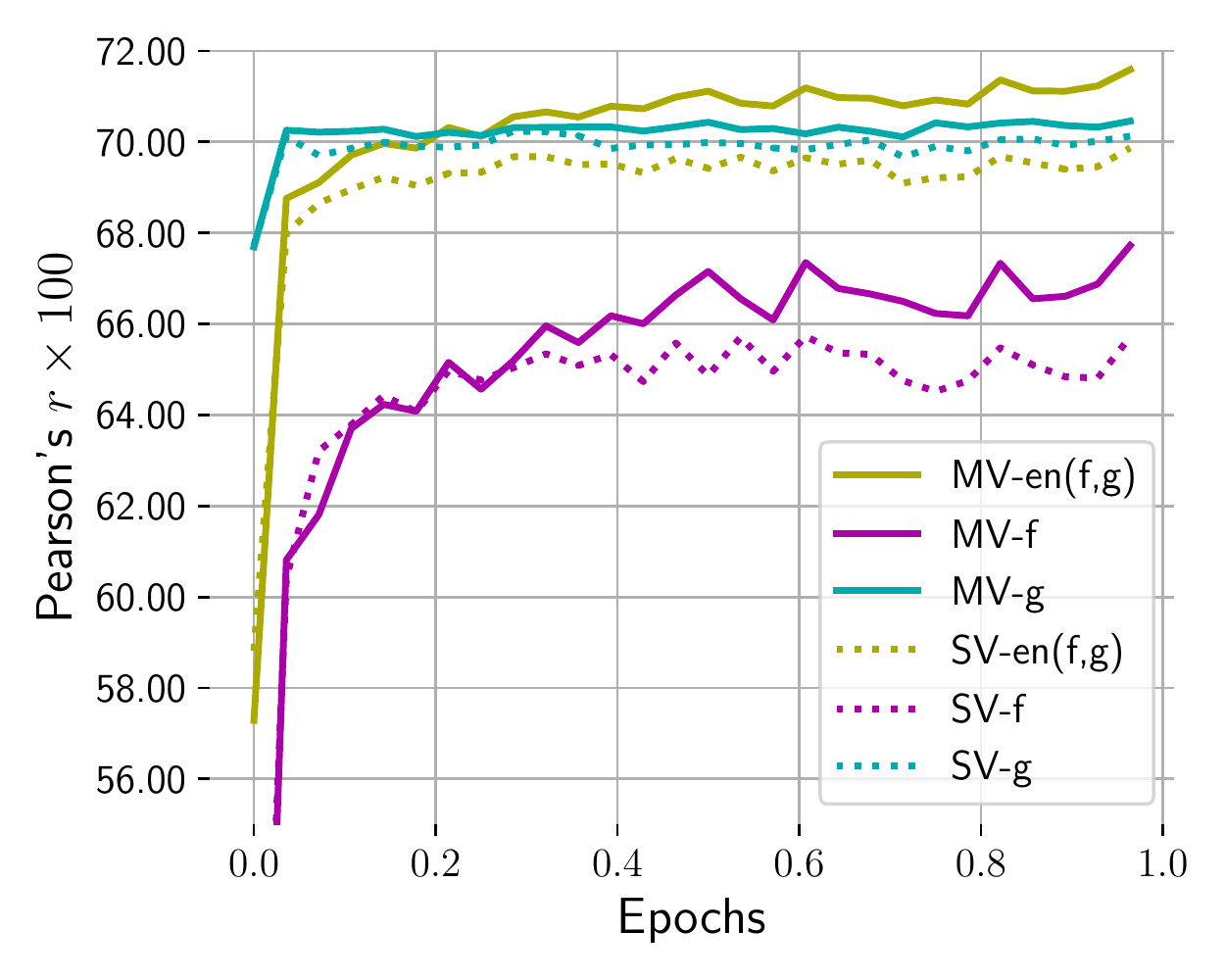}
    \caption{\textbf{Averaged Pearson's correlation score on STS12-16 and SICK14 (six unsupervised evaluation tasks) during learning.} `MV' and `SV' refer to learning with two distinctive views and learning with only one view, respectively. `$f$', `$g$' and `en($f$ and $g$)' refer to the RNN model, the linear model and the ensemble of two encoding functions, individually. \textbf{As shown, our consensus maximisation with two different views improves both $f$ and $g$ on unsupervised tasks compared to their single-view learnt counterparts during learning, and the ensemble of them provides better performance than that of the ensembled counterparts.} (Better viewed in colour.)}
    \label{fig:stsun}
    \end{center}
\vskip -0.2in
\end{figure}

Figure \ref{fig:stsun} illustrates the change of the averaged performance of both our proposed learning method and single-view learning method on unsupervised evaluation tasks. Since the evaluation method in unsupervised tasks is also cosine similarity of the representations of two sentences, which is the same as the learning objective during training, the performance of a model on unsupervised tasks tells us how well the model has learnt so far. In the comparison between our multi-view consensus maximisation and single-view learning, both $f$ and $g$ are improved by learning to align with each other, and the ensemble of them outperforms and ensemble of the single-view learnt counterparts.

\subsection{Skip Connection as a Regulariser}

One could argue that the performance gain of our proposed consensus maximisation learning method with two views against their single-view trained counterparts is due to the fact that the linear model ($g$) serves as a regulariser to help the RNN model ($f$) learn better representations as the widely-used skip connections in both Convolutional Neural Networks \cite{He2016DeepRL,He2016IdentityMI}, multi-layer Recurrent Neural Networks \cite{Wu2016GooglesNM} as well as in the Transformer networks \cite{Vaswani2017AttentionIA}. 

The skip connection directly wires the input to a specific layer up to the output by either directly copying the input or linearly transforming it to match the dimensionality between the input and the output of a layer. It has been shown that the skip connections are effective as a regulariser in stablising the learning process and providing consistent performance gain \cite{He2016IdentityMI}. Here, we aim to study whether the linear model in our consensus maximisation learning method serves as a regulariser to the RNN model.
\begin{align}
    \text{$f$  and  $g$}: a_{ij}=\cos(\vz_i^f + \vz_i^g, \vz_j^f + \vz_j^g)
\end{align}
Instead of having two views learn to agree with each other, this variant adds up the representations from the RNN model and the linear model to serve as a sentence representation. As the linear model in our case is a simple linear projection, it can be regarded as a skip connection that directly connect the input space of the RNN model to its output space and the linear model here serves as a regulariser. The learning objective is still to provide more similar representations for adjacent sentences than those that are not. Results are presented in the fifth section in Table \ref{UMBCmvsv}.

Given the results, the linear model in our consensus maximisation learning method act considerably more than just a regulariser. The performance on the unsupervised evaluation tasks of the RNN model in single-view learning method gets improved by having the linear model as a skip connection (from $57.8$ and $49.7$ to $59.4$), and the ensemble of two views indeed performance better than each one of them which is not the case in the single-view learning method or in the multi-view learning with same architecture. However, our learning method that learns to maximise the agreement between two views (with different architectures) outperforms the variant presented here, which suggests that, \textbf{in our consensus maximisation learning, the linear model has a stronger impact than the skip connection as a regulariser}.

\subsection{Multi-view + Single-view}
A variant that combines both our consensus maximisation learning with two views and single-view learning is proposed, and it is compared against our learning method to show that only maximising between two views is necessary and helpful for each view in learning. The fourth section in the Table \ref{UMBCmvsv} presents the results.
\begin{align}
    a_{ij}&=\cos(\mathbf{z}_i^f,\mathbf{z}_j^g)+\cos(\mathbf{z}_i^g,\mathbf{z}_j^f) & \text{(multi-view)} \nonumber \\
          &+\cos(\mathbf{z}_i^f,\mathbf{z}_j^f)+\cos(\mathbf{z}_i^g,\mathbf{z}_j^g) & \text{(single-view)}
\end{align}
The performance of this variant is slightly worse than our method learnt with only multi-view maximisation, and the ensemble of two views doesn't always provide better results than each one of them on unsupervised evaluation tasks. It is expected as, during learning, we observed that it is much easier to maximise the agreement among representations generated from the same network than to do so between two views, thus single-view maximisation dominates the learning process and eventually leads to degenerated performance.

Since learning from the other view is involved in the learning process, the consensus maximisation learning between two views managed to help each view to learn slightly better than its single-view learnt counterpart. Each view still gets improved on the tasks where the other one excels, although the improvement is not as much as that from learning to align the two views alone.

\section{Detailed Results}

\begin{table*}[!ht]
\fontsize{9}{11}\selectfont
\caption{\textbf{Results on unsupervised evaluation tasks} (Pearson's $r\times 100$) . \textbf{Bold} numbers are the best results among unsupervised transfer models, and \underline{underlined} numbers are the best ones among all models. `D1', `D2' and `D3' refer to three large corpora respectively. `WR' refers to the post-processing step that removes the top principal component. `UP' refers to the post-processing step that is derived from angular based random walk model.}
\begin{center}
\vskip 0.15in
\tabulinesep =_3pt^3pt
\begin{tabu}to \textwidth{@{}| c || c | c | c | c c | c c || c c c || c | c |@{}}
\hline
\multicolumn{13}{|c|}{$^1$\citet{Arora2017ASB};$^2$\citet{Wieting2015FromPD};$^3$\citet{Wieting2018Para};$^4$\citet{Conneau2017SupervisedLO};$^5$\citet{Wieting2018Para};} \\
\multicolumn{13}{|c|}{$^{6-10}$\citet{Agirre2012SemEval2012T6,Agirre2013SEM2S,Agirre2014SemEval2014T1,Agirre2015SemEval2015T2,Agirre2016SemEval2016T1};$^{11}$\citet{Marelli2014ASC};$^{12}$\citet{Mikolov2017AdvancesIP};$^{13}$\citet{Ethayarajh2018UnsupervisedRW}} \\
\hline
\multirow{3}{*}{Task} & \multicolumn{7}{c||}{Un. (Unsupervised) Training} & \multicolumn{3}{c||}{Semi.} & \multicolumn{2}{c|}{Su.} \\
\cline{2-13}
 & \multicolumn{3}{c|}{\textbf{Multi-view}} & \multicolumn{2}{c|}{fastText} & \multicolumn{2}{c||}{GloVe} &\multicolumn{3}{c||}{$^2$PSL} & $^4$Infer & $^3$ParaNMT \\
\cline{2-11}
 & D1 & D2 & D3 & $^{12}$avg & $^1$WR & $^1$tfidf & $^{13}$UP & $^1$avg & $^1$WR & $^{13}$UP & Sent & (concat.) \\
 \hline
$^6$STS12       & 60.9 &  64.0 & 60.7                   & 58.3 & 58.8  & 58.7 &  \textbf{64.9}  & 52.8 & 59.5 & 65.8 & 58.2 & \underline{67.7} \\
$^7$STS13       & 60.1 &  61.7 & 59.9                   & 51.0 & 59.9  & 52.1 &  \textbf{63.6}  & 46.4 & 61.8 & 65.2 & 48.5 & \underline{62.8} \\
$^8$STS14       & 71.5$^\star$ & 73.7 & 70.7            & 65.2 & 69.4  & 63.8 &  \textbf{74.4}  & 59.5 & 73.5 & 78.4 & 67.1 & \underline{76.9} \\
$^9$STS15       & 76.4 &  \textbf{77.2} & 76.5          & 67.7 & 74.2  & 60.6 &  76.1  & 60.0 & 76.3 & 79.0 & 71.1 & \underline{79.8} \\
$^{10}$STS16    & 75.8 &  \textbf{76.7} & 74.8          & 64.3 & 72.4  & -    & -      & -    & -    & - & 71.2 &  \underline{76.8} \\
$^{11}$SICK14   & 74.7 & \textbf{74.9} & 72.8           & 69.8 & 72.3  & 69.4 &  73.0  & 66.4 & 72.9 & 73.5 & 73.4 & - \\
\hline
Average & 69.9 & \textbf{71.4} & 69.2 & 62.7 & 67.8 & - & - & - & - & - & 64.9 & - \\
\hline
\end{tabu}
\end{center}
\label{unsupervised}
\vskip -0.1in
\end{table*}

\subsection{Comparison on unsupervised evaluation tasks}
$\bullet$ \emph{Unsupervised learning}: We selected models with strong results from related work, including fastText, fastText+WR, GloVe+tfidf, and GloVe+UP.

$\bullet$ \emph{Semi-supervised learning}: The word vectors are pretrained on each task \cite{Wieting2015FromPD} without label information, and word vectors are averaged to serve as the vector representation for a given sentence \cite{Arora2017ASB}. The results with both postprocessing steps `WR' and `UP' are reported respectively.

$\bullet$ \emph{Supervised learning}: ParaNMT \cite{Wieting2018Para} is included as a supervised learning method as the data collection requires a neural machine translation system trained in supervised fashion. The InferSent\footnote{The released InferSent \cite{Conneau2017SupervisedLO} model is evaluated with the postprocessing step.} \cite{Conneau2017SupervisedLO} trained on SNLI \cite{Bowman2015ALA} and MultiNLI \cite{Williams2017ABC} is included as well. 

The results are presented in Table \ref{unsupervised}. Since the performance of FastSent \cite{Hill2016LearningDR} and Quick-Thought (QT) \cite{logeswaran2018an} were only evaluated on STS14, we compare to their results in Table \ref{FS_QT}. Our consensus maximisation learning method slightly underperforms GloVe+UP on STS12-14 tasks, and outperforms all other unsupervised learning methods on all other tasks.

\begin{table}[!ht]
\caption{Comparison with FastSent \cite{Hill2016LearningDR} and QT \cite{logeswaran2018an} on STS14 (Pearson's $r\times 100$).}
\fontsize{9}{11}\selectfont
\begin{center}
\vskip 0.15in
\tabulinesep =_3pt^3pt
\begin{tabu}to \textwidth{@{}|c|c|c|c|c|c|c|@{}}
\hline
\multicolumn{2}{|c|}{FastSent} & \multicolumn{2}{c|}{QT} & \multicolumn{3}{c|}{\textbf{Multi-view}} \\
\hline
& +AE & RNN & BOW & D1 & D2 & D3 \\
\hline
61.2 & 59.5 & 49.0 & 65.0 &  71.5$^\star$ & \textbf{73.7} & 70.7 \\
\hline
\end{tabu}
\end{center}
\label{FS_QT}
\vskip -0.1in
\end{table}

\begin{table*}[!ht]
\fontsize{9}{11}\selectfont
\caption{\textbf{Supervised evaluation tasks.} \textbf{Bold} numbers are the best results among unsupervised transfer models, and \underline{underlined} numbers are the best ones among all models. ``\dag'' refers to an ensemble of two models. ``\ddag'' indicates that additional labelled discourse information is required. Our models perform similarly or better than existing methods, but with higher training efficiency.}
\begin{center}
\vskip 0.15in
\tabulinesep =_3pt^3pt
\begin{tabu}to \textwidth{@{}|l| c | c c c | c c c c c c |@{}}
\hline

\multicolumn{11}{|c|}{$^1$\citet{Conneau2017SupervisedLO};$^2$\citet{Arora2017ASB};$^3$\citet{Hill2016LearningDR}; $^4$\citet{Kiros2015SkipThoughtV};} \\
\multicolumn{11}{|c|}{$^5$\citet{Gan2017LearningGS};$^6$\citet{Jernite2017DiscourseBasedOF};$^7$\citet{Nie2017DisSentSR};$^8$\citet{Tai2015ImprovedSR};$^{9}$\citet{Zhao2015SelfAdaptiveHS}}\\
\multicolumn{11}{|c|}{$^{10}$\citet{Le2014DistributedRO};$^{11}$\citet{logeswaran2018an};$^{12}$\citet{Shen2018BaselineNM};$^{13}$\citet{Mikolov2017AdvancesIP}}\\

\Xhline{1.2pt}
Model & Hrs & SICK-R & SICK-E & MRPC & TREC & MR & CR & SUBJ & MPQA & SST \\
\Xhline{1.2pt}
\multicolumn{11}{c}{\emph{Supervised task-dependent training - No transfer learning}} \\
\hline
$^9$AdaSent & - & - & - & - & 92.4 & 83.1 & \underline{86.3} & 95.5 & \underline{93.3} & - \\
$^8$TF-KLD  & - & - & - & \underline{80.4}/\underline{85.9} & - & - & - & - & - & - \\
$^{12}$SWEM-$concat$ & - & - & - & 71.5/81.3 & 91.8 & 78.2 & - & 93.0 & - & 84.3 \\
\hline
\multicolumn{11}{c}{\emph{Supervised training - Transfer learning}} \\
\hline
$^1$InferSent & $<$24 & \underline{88.4} & \underline{86.3} & 76.2/83.1 & 88.2 & 81.1 & 86.3 & 92.4 & 90.2 & 84.6 \\
\hline
\multicolumn{11}{c}{\emph{Unsupervised training with unordered sentences}} \\
\hline
$^{10}$ParagraphVec & 4 & - & -& 72.9/81.1 & 59.4 & 60.2 & 66.9 & 76.3 & 70.7 & - \\
$^2$GloVe+WR & - & 86.0 & 84.6 & - / - & - & - & - & - & - & 82.2 \\
$^{13}$fastText+bow  & - & - & - & 73.4/81.6 & 84.0 & 78.2 & 81.1 & 92.5 & 87.8 & 82.0  \\
$^3$SDAE & 72 & - & - & 73.7/80.7 & 78.4 & 74.6 & 78.0 & 90.8 & 86.9 & -  \\
\hline
\multicolumn{11}{c}{\emph{Unsupervised training with ordered sentences}} \\
\hline
$^3$FastSent & 2 & - & - & 72.2/80.3 & 76.8 & 70.8 & 78.4 & 88.7 & 80.6 & - \\
$^4$Skip-thought & 336 & 85.8 & 82.3 & 73.0/82.0& 92.2 & 76.5 & 80.1 & 93.6 & 87.1 & 82.0 \\
$^5$CNN-LSTM \dag & - & 86.2 & - & 76.5/83.8 & 92.6 & 77.8 & 82.1 & 93.6 & 89.4 & - \\
\hline
$^6$DiscSent \ddag & 8 & - & - & 75.0/ - & 87.2 & - & - & 93.0 & - & - \\
$^7$DisSent \ddag & - & 79.1 & 80.3 & - / - & 84.6 & 82.5 & 80.2 & 92.4 & 89.6 & 82.9 \\
$^{11}$MC-QT  & 11 & 86.8 & - & 76.9/\textbf{84.0} & \underline{\textbf{92.8}} & 80.4 & 85.2 & 93.9 & 89.4 & -\\
\hline
\hline
\textbf{Multi-view} D1 & 3 & 87.9 & 84.8 & 77.1/83.4 & 91.8 & 81.6 & 83.9 & 94.5 & 89.1 & 85.8 \\
\textbf{Multi-view} D2 & 8.5 & 87.8 & 85.2 & 76.8/83.9 & 91.6 & 81.5 & 82.9 & 94.7 & 89.3 & 84.9 \\
\textbf{Multi-view} D3 & 8 & 87.7 & 85.2 & 75.7/82.5 & 89.8 & \underline{\textbf{85.0}} & \textbf{85.7} & \underline{\textbf{95.7}} & \textbf{90.0} & \underline{\textbf{89.6}} \\
\hline
\end{tabu}
\end{center}
\label{supervised}
\vskip -0.1in
\end{table*}

\subsection{Comparison on supervised evaluation tasks}
Our results as well as related results of supervised task-dependent training models, supervised learning models, and unsupervised learning models are presented in Table \ref{supervised}. Note that, for fair comparison, we collect the results of the best single model of MC-QT \cite{logeswaran2018an} trained on BookCorpus. Our consensus maximisation based model is able to provide similar performance as the best unsupervised learning method.

\section{Conclusion} 
We proposed a consensus maximisation based sentence representation learning that combines an RNN-based encoder and an average-on-word-vectors linear encoder and can be efficiently trained within a few hours on a large unlabelled corpus. The experiments were conducted on three large unlabelled corpora, and meaningful comparisons were made to demonstrate the generalisation ability and  transferability of our consensus maximisation learning method and consolidate our claim. The produced sentence representations outperform existing unsupervised transfer methods on unsupervised evaluation tasks, and match the performance of the best unsupervised model on supervised evaluation tasks. 

Based on the results presented in Table \ref{UMBCmvsv}, learning to maximise the agreement between the representations generated from two views helps each view to perform better on downstream tasks compared to their single-view learnt comparison partners. The ensemble of two views in our learning method provides even better performance than each of them, and also better than the ensemble of two single-view learnt counterparts.

Our experimental results support the finding \cite{Hill2016LearningDR} that linear/log-linear models ($g$ in our frameworks) tend to work better on the unsupervised tasks, while RNN-based models ($f$ in our frameworks) generally perform better on the supervised tasks. As presented in our experiments,
consensus maximisation learning method with two distinctive views helps align $f$ and $g$ to produce better individual representations than when they are learned separately. In addition, the ensemble of both views leveraged the advantages of both, and provides rich semantic information of input sentences. Future work should explore the impact of having various encoding architectures and learning under consensus maximisation with multiple views.


\clearpage
\bibliography{example_paper}
\bibliographystyle{icml2019}

\end{document}